\documentclass{article}
\usepackage{iclr2026_conference,times}
\iclrfinalcopy

\usepackage{booktabs}
\usepackage{array}
\usepackage[T1]{fontenc}
\usepackage[utf8]{inputenc}
\usepackage{textcomp}
\usepackage{newunicodechar}
\newunicodechar{ℓ}{\ell}
\usepackage{amsmath}
\usepackage{tcolorbox}

\usepackage{amsmath,amsfonts,bm}

\def\eqref#1{equation~\ref{#1}}
\def\1{\bm{1}}

\DeclareMathAlphabet{\mathsfit}{\encodingdefault}{\sfdefault}{m}{sl}
\SetMathAlphabet{\mathsfit}{bold}{\encodingdefault}{\sfdefault}{bx}{n}

\usepackage{hyperref}
\usepackage{url}
\usepackage{comment}
\usepackage{booktabs}
\usepackage{subcaption}
\usepackage{array}
\usepackage{makecell}
\usepackage{placeins}
\usepackage{float}
\usepackage{tikz}
\usepackage{xcolor}
\usepackage{tabularx}
\usetikzlibrary{positioning, arrows.meta}

\newenvironment{packeditemize}{\begin{list}{$\bullet$}{\setlength{\itemsep}{0.5pt}\addtolength{\labelwidth}{-4pt}\setlength{\leftmargin}{2ex}\setlength{\listparindent}{\parindent}\setlength{\parsep}{1pt}\setlength{\topsep}{2pt}}}{\end{list}}

\title{Test-Time Adaptation via Many-Shot Prompting: Benefits, Limits, and Pitfalls}
\author{
Shubhangi Upasani\thanks{Corresponding author: \texttt{shubhangi.upasani@gmail.com}}\textsuperscript{1},
Chen Wu\textsuperscript{1},
Jay Rainton\textsuperscript{1},
Bo Li\textsuperscript{1},
Urmish Thakker\textsuperscript{1},
\\
Changran Hu \textsuperscript{1},
Qizheng Zhang \textsuperscript{2}\\
\textsuperscript{1}SambaNova Systems, Inc \textsuperscript{2}Stanford University
}

\usepackage{textcomp}
\usepackage{textcomp}
\begin{document}

\maketitle

\begin{abstract}
Test-time adaptation enables large language models (LLMs) to modify their behavior at inference without updating model parameters. A common approach is many-shot prompting, where large numbers of in-context learning (ICL) examples are injected as an input-space test-time update. Although performance can improve as more demonstrations are added, the reliability and limits of this update mechanism remain poorly understood, particularly for open-source models.

We present an empirical study of many-shot prompting across tasks and model backbones, analyzing how performance varies with update magnitude, example ordering, and selection policy. We further study Dynamic and Reinforced ICL as alternative test-time update strategies that control which information is injected and how it constrains model behavior. We find that many-shot prompting is effective for structured tasks where demonstrations provide high information gain, but is highly sensitive to selection strategy and often shows limited benefits for open-ended generation tasks. Overall, we characterize the practical limits of prompt-based test-time adaptation and outline when input-space updates are beneficial versus harmful.
\end{abstract}

\section{Introduction}

Test-time adaptation enables large language models (LLMs) to modify their behavior during inference without a change in model weights~\citep{brown2020language,garg2022can,zhang2025agentic}. 
A widely used but underexplored form of such adaptation is prompt-based input-space updates, where task-relevant information is injected directly into the model’s context. 
With recent advances in long-context architectures~\citep{dao2022flashattention, su2024roformer}, this update mechanism has evolved from few-shot prompting to many-shot prompting, where hundreds or thousands of demonstrations can be provided at test time.

Many-shot prompting can be viewed as an increasingly aggressive test-time update: as more demonstrations are added, the model is exposed to a stronger, task-specific behavioral constraint. 
Prior work~\citep{agarwal2024many} shows that accuracy often improves as the number of in-context examples grows, particularly for structured tasks. 
In this paper, we study many-shot prompting as a controlled study on LLaMA model family, focusing on when and why input-space updates succeed or fail. %Our experiments span extreme label classification, reasoning, and information extraction tasks, as well as domains where many-shot updates are ineffective, such as question answering and machine translation. 
Our key observations from our chosen testbed include the following: 

\begin{packeditemize}
    \item Many-shot prompting can act as an effective test-time update for structured tasks with accuracy improving consistently up to a moderate update magnitude before saturating. 
    \item Reliability of prompt-based test-time updates is highly sensitive to ordering of in-context demonstrations, prompt structure and example selection strategies. 
    \item Test-time updates exhibit clear task-dependent success and failure modes. Although additional examples provide useful adaptation signals for classification and information extraction tasks, they give small improvements for open-ended generation tasks such as machine translation.
    \item Structured updates such as Reinforced ICL exhibit similar saturation behavior as update magnitude increases.
\end{packeditemize}

Together, these findings highlight both the promise and the limits of prompt-based test-time adaptation. 

%However, our empirical evaluation shows that this update mechanism is neither monotonic nor universally generalizable. Increasing the update magnitude can introduce instability, noise, and redundancy, ultimately leading to saturation or even performance degradation.

\section{Background}
\paragraph{Many-Shot Prompting (Update Magnitude):}
With the availability of long-context models, in-context learning (ICL)~\citep{dong2024survey} has extended from a few demonstrations to many-shot prompting, where hundreds or thousands of examples can be included at test time. 
The effectiveness of such updates strongly depends on how the context is constructed; demonstration ordering, diversity, template format and phrasing of instructions play an important role.

\paragraph{Dynamic ICL as Update Policy:}
Dynamic ICL~\citep{rubin2022learning, liu2022makes} constructs prompts on-the-fly by selecting examples for each query based on similarity, heuristics, or task metadata. It is an adaptive update policy that determines the information that is injected into the context.

%Traditional ICL relies on a fixed set of demonstrations for all inputs. In contrast, dynamic ICL constructs prompts on-the-fly by selecting examples for each query based on similarity, heuristics, or task metadata. From a test-time update standpoint, dynamic ICL corresponds to an adaptive update policy that determines which information is injected into the context, trading off relevance and diversity under a fixed context budget. Dynamic ICL swaps out irrelevant examples for more relevant ones, and in practice often leads to meaningful gains in accuracy. 

\paragraph{Reinforced ICL as Update Structure:} 
Reinforced ICL replaces the input-output (IO) demonstrations with reasoning traces, such as chain-of-thought (CoT) examples~\citep{wei2022chain,agarwal2024many}, to guide the behavior of the model at the test time. We treat Reinforced ICL as a structured test-time update. %that constrains the adaptation signal through explicit reasoning patterns. As with many-shot prompting, the effectiveness of this update depends on the number and quality of the demonstrations provided.

\begin{figure}[t]
    \centering
    \includegraphics[width=0.9\linewidth]{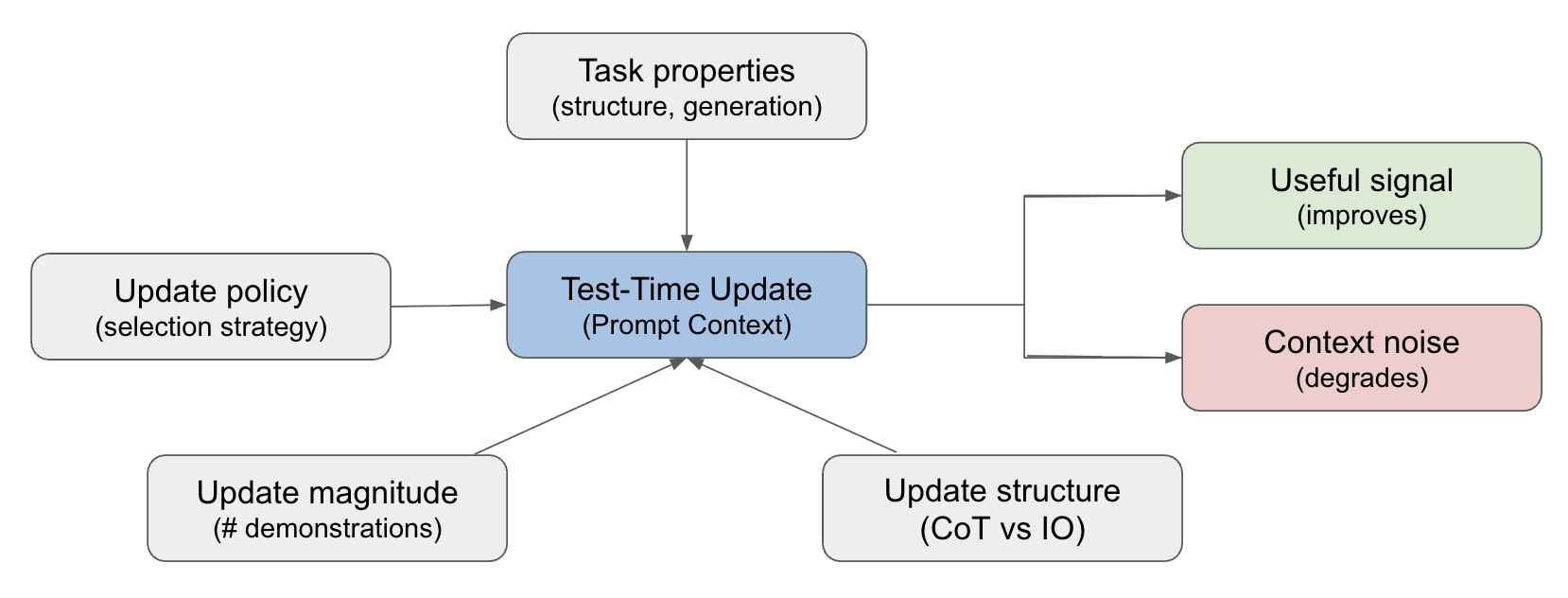}
    \caption{\textbf{A unified view of prompt-based test-time adaptation.}
    Update design determines whether added context provides signal or noise.}
    \label{fig:overview}
\end{figure}

Figure~\ref{fig:overview} summarizes our view of prompt-based adaptation as a test-time update governed by update magnitude, update policy, and update structure, whose interplay determines whether additional context provides useful signal or introduces noise. Throughout this work, we use \textbf{“test-time update”} to refer to input-space adaptation via prompting, 
without implying any parameter-level modification. We give additional background on the above strategies in Appendix \ref{app:icl_prelims}.

\section{Experimental Setup}

\textbf{Tasks and benchmarks:}
We evaluate prompt-based test-time updates across a diverse set of tasks using benchmarks designed to stress long-context inference. For structured classification, we use the Banking77 dataset from LongICLBench~\citep{li2024long} which has 77 classes for intent classification task. This dataset enables controlled evaluation of many-shot prompting under extreme label spaces. To assess success and failure modes, we additionally include tasks from Evaluation Harness~\citep{biderman2024lessons} that span reasoning, information extraction, question answering, and machine translation.

\textbf{Models:}
We evaluate two instruction-tuned backbones of differing capacity: LLaMA-3.1-8B-Instruct \cite{grattafiori2024llama} and LLaMA-3.3-70B-Instruct \cite{meta_llama33_70b} to study how test-time updates scale with model size under identical prompting strategies. Instruction tuning ensures consistent interpretation of task instructions and demonstrations during many-shot adaptation. Base-model ablations, reported in Appendix~\ref{app:base_results}, perform poorly even at high update magnitudes, supporting our focus on instruction-tuned models.

\begin{comment}
\textbf{Test-time update regimes.}
We evaluate three classes of prompt-based test-time updates. (1) \emph{Static many-shot prompting}, where a fixed set of demonstrations is used for all inputs and the update magnitude is varied by the number of examples. (2) \emph{Dynamic ICL}, where demonstrations are selected per query using either random sampling or embedding-based similarity, corresponding to different update policies under a fixed context budget. (3) \emph{Reinforced ICL}, where reasoning traces are provided as demonstrations to induce structured test-time updates.
\end{comment}

\begin{comment}
\textbf{Evaluation protocol:}
For each update regime, we sweep the number of in-context examples to study scaling and saturation behavior. To assess update stability, we repeat experiments across multiple random seeds, varying demonstration ordering and selection where applicable. We report average task accuracy and analyze variance across runs to characterize the reliability of prompt-based test-time updates.
\end{comment}

\begin{comment}
    
\begin{figure}[t]
\centering
\small
\label{fig:ttu_framework}
\begin{tikzpicture}[
     scale=0.75,
    transform shape,
    node distance=1.2cm,
    box/.style={draw, rounded corners, align=center, minimum width=1.6cm, minimum height=0.4cm},
    arrow/.style={->, thick}
]

% Central node
\node[box] (update) {Test-Time Update\\(Prompt Context)};

% Surrounding nodes
\node[box, above=of update] (task) {Task Properties\\(structure, novelty)};
\node[box, left=of update] (policy) {Update Policy\\(selection strategy)};
\node[box, right=of update] (structure) {Update Structure\\(IO vs CoT)};
\node[box, below=of update] (magnitude) {Update Magnitude\\(\# demonstrations)};

% Arrows
\draw[arrow] (policy) -- (update);
\draw[arrow] (structure) -- (update);
\draw[arrow] (magnitude) -- (update);
\draw[arrow] (task) -- (update);

\end{tikzpicture}
\caption{A unified view of prompt-based test-time adaptation. Prompting acts as an input-space update governed by task properties, update magnitude, update policy, and update structure.}
\label{fig:ttu_framework}
\end{figure}
\end{comment}

\section{Results: Benefits And Limits of Test-Time Updates}

\begin{comment}
\subsection{Scaling with Update Magnitude}

We first evaluate how the effectiveness of prompt-based test-time updates varies with update magnitude, defined by the number of in-context demonstrations. For Banking77 with $C{=}77$ classes, we construct prompts with $N{=}n\times C$ total demonstrations (e.g., $n{=}1 \Rightarrow N{=}77$, $n{=}5 \Rightarrow N{=}385$), and vary how those $N$ examples are chosen. We use this method to structure the prompt to give appropriate coverage for each class in the dataset. To assess update stability, we repeat experiments across multiple random seeds, varying demonstration ordering and selection where applicable. We report the average accuracy of the task and analyze the variance between runs.

As shown in Table~\ref{tab:banking77_updates_table_1}, accuracy increases steadily from 68\% (1-shot) to $\sim$82\% (70 shots), after which performance slightly plateaus. 
\end{comment}

\subsection{More Context Helps—Until It Doesn’t}

We evaluate how the effectiveness of prompt-based test-time updates varies with update magnitude, defined by the total number of in-context demonstrations. For Banking77 with $C=77$ classes, we use a \emph{per-class} shot formulation: each prompt contains $n$ demonstrations per class, resulting in $N = n \times C$ total demonstrations (e.g., $n=1 \Rightarrow N=77$, $n=5 \Rightarrow N=385$). %This construction ensures uniform class coverage in the prompt and avoids bias toward frequent labels.

Unless otherwise stated, we refer to $n$ as the \emph{per-class shot count} and $N$ as the \emph{total number of demonstrations}. 
%To assess update stability, we repeat experiments across multiple random seeds, varying demonstration ordering and selection, and report mean accuracy and standard deviation. As shown in Table~\ref{tab:banking77_updates_table_1}, accuracy increases steadily from 68\% with $n=1$ per-class shot ($N=77$ total demonstrations) to approximately 82\% at $n=70$ ($N=5390$), after which performance plateaus. These results indicate a clear saturation regime: moderate test-time updates provide meaningful adaptation, while more aggressive updates yield diminishing returns. We stop at N = 5390-shots as we reach the maximum context length of 128K for Llama models. (refer Appendix \ref{app:context_length_plot})
As shown in Fig.~\ref{fig:ttu_plots}(a), as the update magnitude increases for LLaMA-3.1-8B-Instruct, accuracy improves steadily, after which it reaches a plateau at approximately 50–70 shots per class. Beyond this point, additional demonstrations provide diminishing returns. This behavior indicates a clear saturation regime: moderate input-space updates introduce useful task-specific signal, while aggressive updates yield diminishing further gains \cite{agarwal2024many}. We hypothesize that the saturation observed in many-shot prompting arises from the limited representational capacity of attention-based conditioning. Demonstrations influence predictions through their key–value representations in the context, but the resulting update is a weighted aggregation. Once sufficient demonstrations are present to identify the task pattern, additional examples contribute largely redundant representations, causing marginal gains to diminish and, in very long prompts, attention competition can even lead to performance degradation.

To assess update reliability, we repeat each n-shot setting using 10 random demonstration orderings. Fig.~\ref{fig:ttu_plots}(a) shows that accuracy varies by 2-3 \% across shuffles, demonstrating that many-shot prompting remains sensitive to demonstration ordering due to positional and contextual bias. Averaging over multiple random orders yields more stable results. We explain more experimental details in Appendix \ref{app:results_section_1}. 

\subsection{Update policy matters—relevance helps early, diversity helps at scale}

Next, we study how update policy affects test-time adaptation under a fixed context budget in Fig.~\ref{fig:ttu_plots}(b) for LLaMA-3.1-8B-Instruct. Across all comparisons, we fix the total number of demonstrations $N$ and vary only the update policy used to select these. We ensure prompt template consistency and prevent including query or same-text duplicates in selected examples.

\textbf{Label-wise vs. Cross-label:} We construct prompts using a per-class formulation with $n$ demonstrations per class, resulting in $N = n \times C$ total demonstrations for Banking77. In label-wise selection, we enforce balanced coverage by selecting exactly $n$ demonstrations per class. In cross-label selection, we select $N$ demonstrations from the entire dataset, so label frequencies can be uneven. 

\textbf{Random vs. similarity:} Within either grouping, random selection samples demonstrations uniformly across the entire dataset, while similarity selection retrieves nearest neighbors to the query in embedding space. More details given in Appendix \ref{app:dynamic_icl} and \ref{app:sim_ret}.

Comparing these policies, cross-label selection consistently outperforms label-wise selection, suggesting that enforcing per-label balance can reduce useful diversity by over representing redundant examples. Cross-label benefits from exposure to greater contextual diversity thereby improving generalization, label-balance forces many low information examples which is not very helpful. Cross-label similarity-based selection is strong at small update magnitudes (high relevance), but degrades as $N$ grows, whereas cross-label random selection scales more robustly with larger $N$ (higher diversity). This reflects a bias-variance tradeoff in test-time updates: relevance-focused (similarity-based) policies help early but can over concentrate the context as the update becomes aggressive; diversity-focused (random) policies scale better. Our best setting is ($n=1 \Rightarrow N=77$) for cross label similarity. This maximizes relevance per demonstration per class label, yielding strong task adaptation before redundancy degrades performance at larger scale. Since cross-label sampling does not enforce uniform class coverage, improvements in overall accuracy may partially reflect gains on frequent classes; future work could analyze per-class accuracy or macro-F1 to better understand the impact on tail classes. Additional results included in Appendix \ref{app:dynamic_icl}.

\subsection{Larger models benefit earlier, smaller models catch up}

\begin{figure}[t]
\centering
\includegraphics[width=1.0\linewidth]{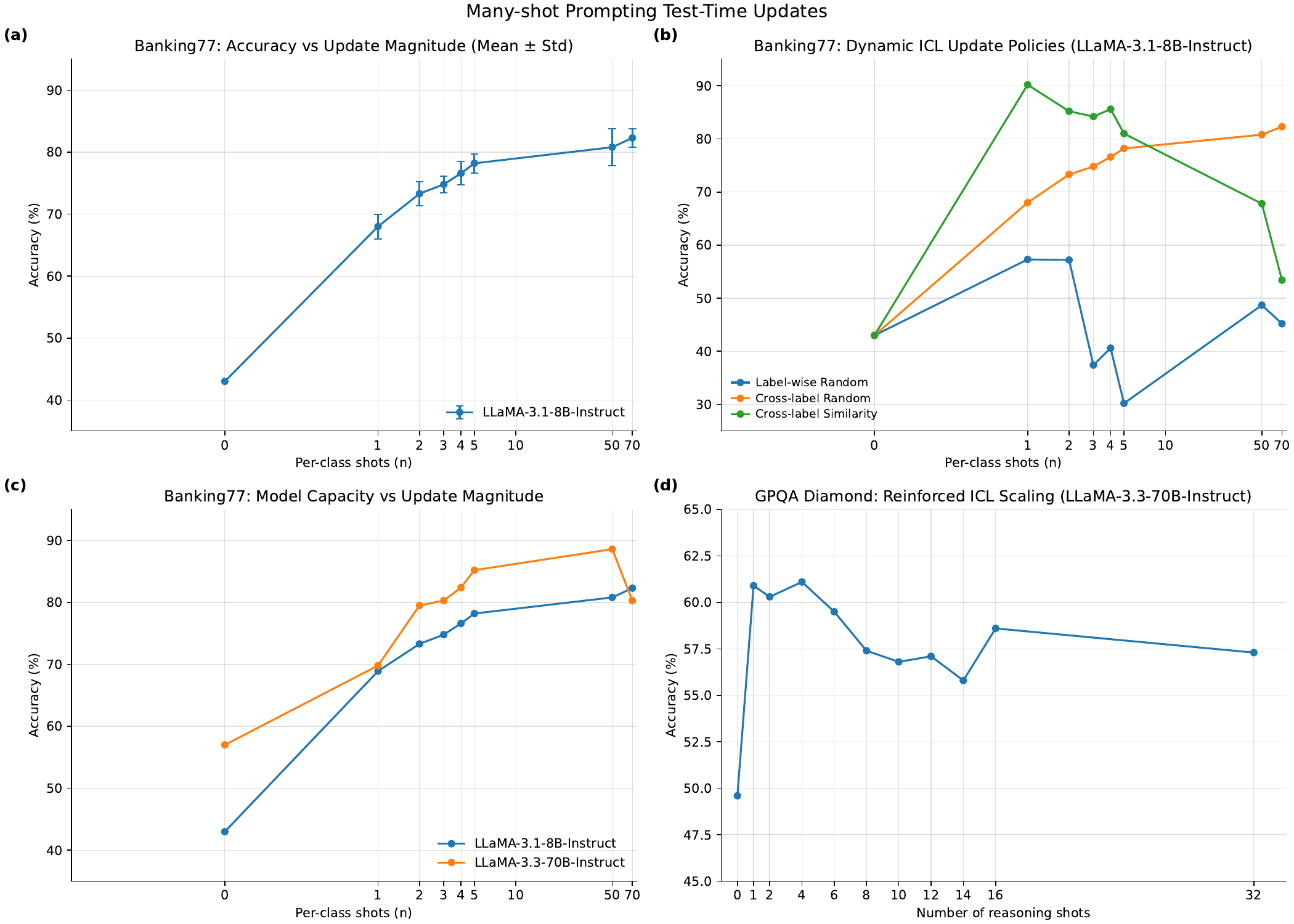}
\caption{
Scaling behavior of prompt-based test-time updates.
(a) Many-shot accuracy vs. update magnitude
(b) Dynamic ICL update policies
(c) Model capacity effects
(d) Reinforced ICL scaling 
}
\label{fig:ttu_plots}
\end{figure}

We further compare test-time update behavior across model sizes using LLaMA-3.1-8B-Instruct and LLaMA-3.3-70B-Instruct in Fig.~\ref{fig:ttu_plots}(c) under identical settings for cross label random selection on Banking 77. 
At small to moderate update magnitudes, LLaMA-70B consistently outperforms the smaller model, indicating that higher capacity models can more effectively exploit diverse in-context supervision. As the update magnitude increases further, the performance gap narrows and the 8B model catches up, suggesting that sufficiently large prompt can partially compensate for limited model capacity. Notably, the 70B model exhibits a performance drop at the largest update magnitude, consistent with over-conditioning. In contrast, the smaller model remains in a signal-accumulation regime and is less sensitive to this effect.

\begin{comment}

\begin{table}[t]
\centering
\caption{Representative success/failure modes of many-shot prompting (accuracy in \%)}
\label{tab:task_success_failure}
\footnotesize
\setlength{\tabcolsep}{3pt}
\renewcommand{\arraystretch}{0.88}
\begin{tabular}{l c c c}
\toprule
Task & 4 & 16 & 32 \\
\midrule
\textbf{Info-heavy / Domain} \\
FDA    & 86.8 & 89.4 & 89.7 \\
SWDE   & 92.9 & 95.1 & 96.4 \\
MEQSum & 3.4  & 6.8  & 11.2 \\
\midrule
\textbf{Constrained reasoning} \\
ARC-Challenge & 39.3 & 39.3 & 39.3 \\
GSM8K         & 94.7 & 94.8 & 94.8 \\
\midrule
\textbf{Open-ended generation} \\
WMT16 En-De & 37.0 & 37.4 & 37.5 \\
GPQA (Gen)  & 48.1 & 46.5 & 44.2 \\
\bottomrule
\end{tabular}
\vspace{-2mm}
\end{table}
\end{comment}

\subsection{Reinforced ICL exhibits early gains and rapid saturation}

We next study Reinforced ICL as a form of structured test-time update, where chain-of-thought (CoT) based reasoning traces \cite{agarwal2024many, wei2022chain} are provided as demonstrations instead of direct input-output pairs. We delineate the data generation pipeline to obtain these samples in the Appendix \ref{app:data_gen}. 
%Using GPQA Diamond subset with LLaMA-3.3-70B-Instruct, we sweep the number of reasoning demonstrations to examine scaling behavior.
Using GPQA Diamond \cite{rein2024gpqa} subset with LLaMA-3.3-70B-Instruct, we show in Fig.~\ref{fig:ttu_plots}(d), Reinforced ICL improves performance as the magnitude of the update increases to 4 demonstrations. Beyond this point, accuracy plateaus and degrades. A plausible explanation is that early demonstrations provide a strong inductive
bias, leading to rapid gains
with only a few examples. As the number of reasoning traces increases, attention is increasingly divided across long chains-of-thought. This competition for attention reduces the effective influence of any single trace, causing performance
to plateau or degrade despite additional context \cite{liu2024lost}.

\subsection{Task structure determines the effectiveness of test-time updates}

\begin{comment}
    
Finally, we analyze when prompt-based test-time updates are effective across tasks (Table~\ref{tab:task_success_failure_full} in Appendix~\ref{app:eval_harness}) using LLama-3.3-70B-Instruct. We find that many-shot prompting consistently improves performance on structure-heavy tasks with constrained outputs, including structured reasoning (e.g., DROP) and information extraction benchmarks (e.g., FDA, SWDE). In these settings, additional demonstrations provide high information gain by capturing relevant input-output patterns. In contrast, gains are limited for tasks with highly constrained answer spaces (e.g., ARC, GSM8K), where a small number of demonstrations is often sufficient to specify task behavior. We observe little to no benefit for open-ended generation tasks such as question answering (e.g., GPQA-Multi Choice) and machine translation (e.g., WMT), where additional demonstrations largely reiterate patterns already captured during pretraining,
\end{comment}

Finally, we analyze when prompt-based test-time updates are effective across tasks (Table~\ref{tab:task_success_failure_full} in Appendix~\ref{app:eval_harness}) using LLaMA-3.3-70B-Instruct. We find that many-shot prompting consistently improves performance on structure-heavy tasks with constrained outputs, including structured reasoning (e.g., DROP \cite{dua2019drop}) and information extraction benchmarks (e.g., FDA, SWDE \cite{hao2011swde}). In these settings, additional demonstrations provide high information gain by capturing relevant patterns. For tasks with constrained outputs (e.g., ARC-Challenge \cite{clark2018think} and GSM8K \cite{cobbe2021training}), performance improves sharply with a small number of demonstrations but quickly saturates, indicating that only limited contextual supervision is required to specify task behavior. In contrast, GPQA (multiple choice) exhibits only modest gains at small update magnitudes. Finally, open-ended generation tasks such as machine translation (WMT16 \cite{bojar-etal-2016-findings}) show consistent but small improvements with additional context, indicating test-time updates offer limited benefits when task structures are already well captured during pretraining.

%In these settings, additional demonstrations provide high information gain by clarifying alignment between input patterns and structured output formats. , and additional examples become redundant.

%More broadly, our results suggest that prompt-based test-time updates are most effective when demonstrations reduce ambiguity in output structure or decision boundaries, such as in tasks with high label entropy (eg: Banking 77) or non-trivial extraction schemas (eg: FDE, DROP). Conversely,  increasing contextual noise rather than introducing new adaptation signal.

\begin{comment}
    
We conclude our experimental results with the following key takeaways: 

\begin{packeditemize}
    \item Many-shot prompting acts as a controllable test-time update, not a free lunch.
    \item Update magnitude exhibits saturation and instability beyond moderate scales (50-70 n-shots).
    \item Update policy (selection) and model backbone size matters as much as update size.
    \item Structured updates (Reinforced ICL) follow similar saturation dynamics.
    \item Structured tasks with high label ambiguity benefit the most from many shot prompting.
    \item Open-ended generation does not benefit from aggressive test-time updates.
\end{packeditemize}
\end{comment}

\section{Conclusion}

Our study highlights both the promise and the practical limits of many-shot prompting as a form of test-time adaptation. We show that increasing the number of in-context demonstrations can serve as an effective input-space update for structured tasks. Across tasks and models, performance is highly sensitive to update configuration, including demonstration ordering and selection policy. Dynamic ICL with cross-label selection produces more robust adaptation than static or label-wise prompting, underscoring the importance of update policy design at test time. At the same time, scaling benefits saturate beyond moderate update magnitudes. Finally, we find that many-shot prompting offers limited benefit for open-ended generation tasks, but works well for structured tasks. Taken together, our results suggest that prompt-based test-time updates are most effective when demonstrations inject novel, task-specific information, and that careful control over update magnitude, structure, and policy is essential for reliable deployment. While our experiments focus on a single benchmark, the observed behaviors should be interpreted as preliminary empirical findings, and future work should validate whether these trends generalize across additional datasets and task domains.

% \newpage



\bibliography{iclr2026_conference}

@inproceedings{dong2024survey,
  title={A survey on in-context learning},
  author={Dong, Qingxiu and Li, Lei and Dai, Damai and Zheng, Ce and Ma, Jingyuan and Li, Rui and Xia, Heming and Xu, Jingjing and Wu, Zhiyong and Chang, Baobao and others},
  booktitle={Proceedings of the 2024 conference on empirical methods in natural language processing},
  pages={1107--1128},
  year={2024}
}

@article{brown2020language,
  title={Language models are few-shot learners},
  author={Brown, Tom and Mann, Benjamin and Ryder, Nick and Subbiah, Melanie and Kaplan, Jared D and Dhariwal, Prafulla and Neelakantan, Arvind and Shyam, Pranav and Sastry, Girish and Askell, Amanda and others},
  journal={Advances in neural information processing systems},
  volume={33},
  pages={1877--1901},
  year={2020}
}

@article{garg2022can,
  title={What can transformers learn in-context? a case study of simple function classes},
  author={Garg, Shivam and Tsipras, Dimitris and Liang, Percy S and Valiant, Gregory},
  journal={Advances in neural information processing systems},
  volume={35},
  pages={30583--30598},
  year={2022}
}

@article{dao2022flashattention,
  title={Flashattention: Fast and memory-efficient exact attention with io-awareness},
  author={Dao, Tri and Fu, Dan and Ermon, Stefano and Rudra, Atri and R{\'e}, Christopher},
  journal={Advances in neural information processing systems},
  volume={35},
  pages={16344--16359},
  year={2022}
}

@article{su2024roformer,
  title={Roformer: Enhanced transformer with rotary position embedding},
  author={Su, Jianlin and Ahmed, Murtadha and Lu, Yu and Pan, Shengfeng and Bo, Wen and Liu, Yunfeng},
  journal={Neurocomputing},
  volume={568},
  pages={127063},
  year={2024},
  publisher={Elsevier}
}

@article{agarwal2024many,
  title={Many-shot in-context learning},
  author={Agarwal, Rishabh and Singh, Avi and Zhang, Lei and Bohnet, Bernd and Rosias, Luis and Chan, Stephanie and Zhang, Biao and Anand, Ankesh and Abbas, Zaheer and Nova, Azade and others},
  journal={Advances in Neural Information Processing Systems},
  volume={37},
  pages={76930--76966},
  year={2024}
}

@article{li2024long,
  title={Long-context llms struggle with long in-context learning},
  author={Li, Tianle and Zhang, Ge and Do, Quy Duc and Yue, Xiang and Chen, Wenhu},
  journal={arXiv preprint arXiv:2404.02060},
  year={2024}
}

@article{biderman2024lessons,
  title={Lessons from the trenches on reproducible evaluation of language models},
  author={Biderman, Stella and Schoelkopf, Hailey and Sutawika, Lintang and Gao, Leo and Tow, Jonathan and Abbasi, Baber and Aji, Alham Fikri and Ammanamanchi, Pawan Sasanka and Black, Sidney and Clive, Jordan and others},
  journal={arXiv preprint arXiv:2405.14782},
  year={2024}
}

@inproceedings{rubin2022learning,
  title={Learning to retrieve prompts for in-context learning},
  author={Rubin, Ohad and Herzig, Jonathan and Berant, Jonathan},
  booktitle={Proceedings of the 2022 conference of the North American chapter of the association for computational linguistics: human language technologies},
  pages={2655--2671},
  year={2022}
}

@inproceedings{liu2022makes,
  title={What makes good in-context examples for GPT-3?},
  author={Liu, Jiachang and Shen, Dinghan and Zhang, Yizhe and Dolan, William B and Carin, Lawrence and Chen, Weizhu},
  booktitle={Proceedings of Deep Learning Inside Out (DeeLIO 2022): The 3rd workshop on knowledge extraction and integration for deep learning architectures},
  pages={100--114},
  year={2022}
}

@article{wei2022chain,
  title={Chain-of-thought prompting elicits reasoning in large language models},
  author={Wei, Jason and Wang, Xuezhi and Schuurmans, Dale and Bosma, Maarten and Xia, Fei and Chi, Ed and Le, Quoc V and Zhou, Denny and others},
  journal={Advances in neural information processing systems},
  volume={35},
  pages={24824--24837},
  year={2022}
}

@article{zhang2025agentic,
  title={Agentic context engineering: Evolving contexts for self-improving language models},
  author={Zhang, Qizheng and Hu, Changran and Upasani, Shubhangi and Ma, Boyuan and Hong, Fenglu and Kamanuru, Vamsidhar and Rainton, Jay and Wu, Chen and Ji, Mengmeng and Li, Hanchen and others},
  journal={arXiv preprint arXiv:2510.04618},
  year={2025}
}

@misc{camelai_datagen,
  title        = {Data Generation Module — Camel AI Documentation},
  author       = {Camel AI},
  year         = {2025},
  howpublished = {\url{https://docs.camel-ai.org/key_modules/datagen}},
  note         = {Accessed: 2026-01-XX},
}

@article{liu2024lost,
  title={Lost in the middle: How language models use long contexts},
  author={Liu, Nelson F and Lin, Kevin and Hewitt, John and Paranjape, Ashwin and Bevilacqua, Michele and Petroni, Fabio and Liang, Percy},
  journal={Transactions of the association for computational linguistics},
  volume={12},
  pages={157--173},
  year={2024}
}

@article{dua2019drop,
  title={DROP: A reading comprehension benchmark requiring discrete reasoning over paragraphs},
  author={Dua, Dheeru and Wang, Yizhong and Dasigi, Pradeep and Stanovsky, Gabriel and Singh, Sameer and Gardner, Matt},
  journal={arXiv preprint arXiv:1903.00161},
  year={2019}
}

@article{clark2018think,
  title={Think you have solved question answering? try arc, the ai2 reasoning challenge},
  author={Clark, Peter and Cowhey, Isaac and Etzioni, Oren and Khot, Tushar and Sabharwal, Ashish and Schoenick, Carissa and Tafjord, Oyvind},
  journal={arXiv preprint arXiv:1803.05457},
  year={2018}
}

@article{cobbe2021training,
  title={Training verifiers to solve math word problems},
  author={Cobbe, Karl and Kosaraju, Vineet and Bavarian, Mohammad and Chen, Mark and Jun, Heewoo and Kaiser, Lukasz and Plappert, Matthias and Tworek, Jerry and Hilton, Jacob and Nakano, Reiichiro and others},
  journal={arXiv preprint arXiv:2110.14168},
  year={2021}
}

@inproceedings{bojar-etal-2016-findings,
    title = "Findings of the 2016 Conference on Machine Translation ({WMT}16)",
    author = "Bojar, Ond{\v{r}}ej  and
      Chatterjee, Rajen  and
      Federmann, Christian  and
      Graham, Yvette  and
      Haddow, Barry  and
      Huck, Matthias  and
      Yepes, Antonio Jimeno  and
      Koehn, Philipp  and
      Logacheva, Varvara  and
      Monz, Christof  and
      Negri, Matteo  and
      Neveol, Aurelie  and
      Neves, Mariana  and
      Popel, Martin  and
      Specia, Lucia  and
      Turchi, Marco  and
      Vitas, Du{\v{s}}ko",
    booktitle = "Proceedings of the First Conference on Machine Translation (WMT16)",
    month = aug,
    year = "2016",
    address = "Berlin, Germany",
    publisher = "Association for Computational Linguistics",
    url = "https://aclanthology.org",
    pages = "131--198"
}

@inproceedings{rein2024gpqa,
  title={Gpqa: A graduate-level google-proof q\&a benchmark},
  author={Rein, David and Hou, Betty Li and Stickland, Asa Cooper and Petty, Jackson and Pang, Richard Yuanzhe and Dirani, Julien and Michael, Julian and Bowman, Samuel R},
  booktitle={First Conference on Language Modeling},
  year={2024}
}

@inproceedings{hao2011swde,
  title={Automatic Web Data Extraction from Large Websites},
  author={Hao, Ming and Madhavan, Jyotish and Szekely, Pedro and Halevy, Alon Y.},
  booktitle={Proceedings of the VLDB Endowment},
  volume={4},
  number={12},
  pages={1326–1337},
  year={2011},
  organization={VLDB}
}

@misc{openai_gptoss120b,
  title        = {GPT-OSS-120B Model Documentation},
  author       = {{OpenAI}},
  year         = {2024},
  note         = {Accessed: 2026-01-XX},
  url          = {https://platform.openai.com/docs/models/gpt-oss-120b}
}

@article{grattafiori2024llama,
  title={The llama 3 herd of models},
  author={Grattafiori, Aaron and Dubey, Abhimanyu and Jauhri, Abhinav and Pandey, Abhinav and Kadian, Abhishek and Al-Dahle, Ahmad and Letman, Aiesha and Mathur, Akhil and Schelten, Alan and Vaughan, Alex and others},
  journal={arXiv preprint arXiv:2407.21783},
  year={2024}
}

@misc{meta_llama33_70b,
  title        = {{LLaMA-3.3-70B-Instruct} Model Card},
  author       = {{Meta AI}},
  year         = {2024},
  note         = {Accessed: 2026-01-XX},
  url          = {https://github.com/meta-llama/llama-models/blob/main/models/llama3_3/MODEL_CARD.md}
}
\bibliographystyle{iclr2026_conference}

\newpage

\appendix
\section{Appendix}

\subsection{Additional Background on Prompt-Based Test-Time Adaptation}
\label{app:icl_prelims}

We provide additional context on how prompt-based adaptation mechanisms differ in the strength, structure, and selectivity of the update signal they introduce at test time.

\textbf{In-context learning as test-time updates:}
In-context learning (ICL) refers to the ability of large language models to adapt their behavior at inference by conditioning on task-specific examples provided in the input, without modifying model parameters. From a test-time perspective, ICL constitutes an input-space update in which task information is injected directly into the context, enabling adaptation without training.

\textbf{Update magnitude in many-shot prompting:}
Many-shot prompting increases the strength of test-time update by injecting a larger volume of task-specific signal into the context. While additional demonstrations can improve performance by reinforcing task patterns, they also increase redundancy and noise, making update effectiveness sensitive to prompt composition. In practice, ordering, diversity, and formatting choices determine whether additional examples provide new information or overwhelm the model with repeated context.

\textbf{Selectivity in Dynamic ICL:}
Dynamic ICL controls which information is injected into the prompt under a fixed context budget. By selecting demonstrations per query, dynamic policies trade off relevance and diversity. Similarity-based selection prioritizes relevance to the input, while broader sampling increases coverage of the task distribution. This selectivity distinguishes dynamic ICL from static prompting, where the same update is applied uniformly across inputs.

\textbf{Structure in Reinforced ICL:}
Reinforced ICL constrains test-time adaptation through structured reasoning traces rather than direct input–output mappings. By exposing intermediate reasoning steps, the update signal emphasizes how to solve a task instead of what answer to produce. As a result, performance depends not only on the number of demonstrations but also on the consistency and quality of the reasoning patterns being demonstrated.

\subsection{Additional Experimental Details for Update Magnitude Experiments}
\label{app:results_section_1}

For the results shown in Fig.~\ref{fig:ttu_plots}(a) on Banking77, we vary the per-class
shot count \(n\). The total number of demonstrations is given by
\(N = n \times C\), where \(C = 77\) is the number of classes.
Table~\ref{tab:n_vs_N} illustrates how \(N\) scales with \(n\).
We stop at \(N = 5390\) demonstrations, corresponding to the maximum supported
context length of 128k tokens for LLaMA models.

To assess update stability, we repeat each setting across multiple random seeds,
varying both demonstration ordering and selection, and report mean accuracy
along with standard deviation.

\begin{table}[t]
\centering
\caption{Comparison of per-class shots with total demonstrations in input prompt for Banking 77}
\label{tab:n_vs_N}
\begin{tabular}{c c}
\toprule
\makecell{\textbf{$n$}\\\textbf{per-class shots}} &
\makecell{\textbf{$N$}\\\textbf{Total demonstrations}} \\
\midrule
1  & 77   \\
2  & 154  \\
3  & 231  \\
4  & 308  \\
5  & 385  \\
50 & 3850 \\
70 & 5390 \\
\bottomrule
\end{tabular}
\end{table}

%As shown in Table~\ref{tab:banking77_updates_table_1}, accuracy increases steadily from 68\% with $n=1$ per-class shot ($N=77$ total demonstrations) to approximately 82\% at $n=70$ ($N=5390$), after which performance plateaus. These results indicate a clear saturation regime: moderate test-time updates provide meaningful adaptation, while more aggressive updates yield diminishing returns. We stop at N = 5390-shots as we reach the maximum context length of 128K for Llama models

\subsection{Dynamic ICL selection strategies}
\label{app:dynamic_icl}

In Dynamic ICL, demonstrations are selected adaptively for each input query rather than using a fixed prompt. We instantiate Dynamic ICL by combining two axes: (i) grouping constraint and (ii) retrieval rule. For a dataset with $C$ classes and per-class shot parameter $n$, the prompt contains $N=n\times C$ demonstrations.

\textbf{Grouping constraint:}
\begin{itemize}
    \item \textbf{Label-wise:} enforce exactly $n$ demonstrations from each class ($N=nC$).
    \item \textbf{Cross-label:} select $N$ demonstrations globally; class counts are unconstrained.
\end{itemize}

\textbf{Retrieval rule:}
\begin{itemize}
    \item \textbf{Random:} uniform sampling from the entire dataset. The results of random selection are averaged across multiple seeds to ensure consistency.
    \item \textbf{Similarity:} top-$N$ retrieval by embedding similarity to the query.
\end{itemize}

Combining the above two regimes, we get four strategies used in our experiments:
\begin{itemize}
    \item Cross-label Random Selection
    \item Label-wise Random Selection
    \item Cross-label Similarity Selection
    \item Label-wise Similarity Selection
\end{itemize}

\begin{figure}[t]
    \centering
    \includegraphics[width=0.9\linewidth]{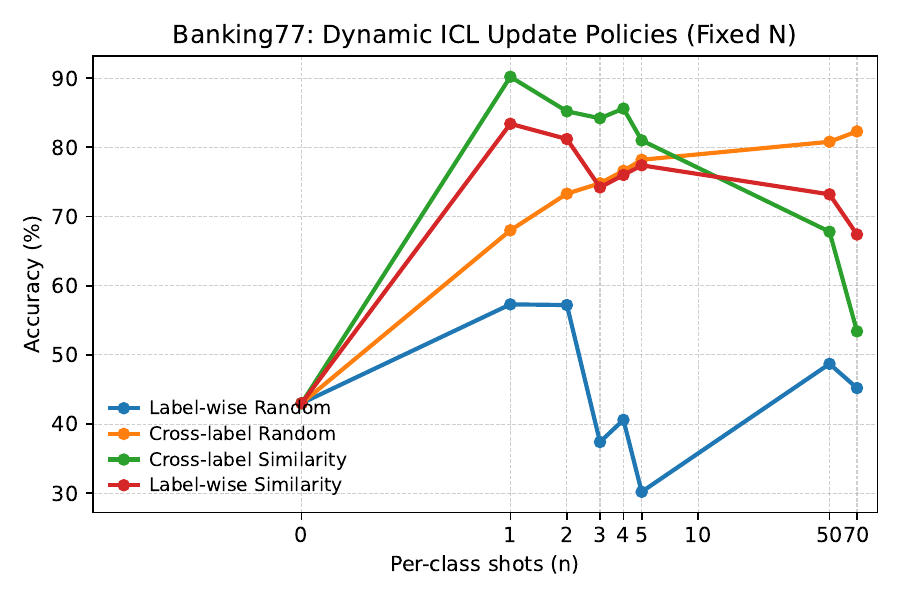}
    \caption{Dynamic ICL Selection Strategies on Banking 77}
    \label{fig:all_strats}
\end{figure}

This taxonomy allows us to isolate the effects of relevance, diversity, and label balance on test-time adaptation behavior. We make sure that the retrieved examples do not have the query and same-text duplicates, to prevent leakage while evaluating the above selection strategies. We also ensure to keep the instruction and prompt templates constant across all runs. We add label-wise similarity as an additional comparison to Fig.~\ref{fig:ttu_plots}(d) and show the full set of results in Fig. \ref{fig:all_strats} for LLaMA-3.1-8B-Instruct.

%We further report additional comparisons between label-wise random and label-wise similarity based selection in Table \ref{tab:labelwise_random_vs_similarity}. 

\begin{comment}

\begin{table}[t]%[H]
\centering
\caption{Label-wise Dynamic ICL on Banking77 (accuracy in \%). Prompts contain $N$ total demonstrations, constructed using $n$ shots per class 
($N = n \times C$, $C{=}77$).}
\label{tab:labelwise_random_vs_similarity}
\begin{tabular}{c c c}
\toprule
\makecell{Per-class\\shots $n$} & Label-wise Random & Label-wise Similarity \\
\midrule
0 (Baseline) & 43.0 & 43.0 \\
1  & 57.3 & 83.4 \\
2  & 57.2 & 81.2 \\
3  & 37.4 & 74.2 \\
4  & 40.6 & 76 \\
5  & 30.2 & 77.4 \\
50 & 48.7 & 73.2 \\
70 & 45.2 & 67.4 \\
\bottomrule
\end{tabular}
\end{table}
\end{comment}

We observe that label-wise similarity selection consistently outperforms label-wise random selection across all update magnitudes. This indicates that when label balance is enforced, selecting demonstrations that are semantically relevant to the query provides a stronger adaptation signal than random sampling within each class.

However, performance under label-wise similarity selection still degrades as the number of demonstrations increases (e.g., from 83.4\% at $n$ = 1 per-class shot to 67.4\% at $n$ = 70 per-class shot). This suggests that enforcing per-label balance limits the effective diversity of the prompt: as the update becomes more aggressive, additional demonstrations become increasingly redundant, even when selected by similarity. Label-wise similarity outperforms cross-label similarity at higher n-shots because restricting retrieval to a fixed label avoids conflicting supervision: although examples become increasingly redundant, they remain label-consistent as update magnitude increases. Label-wise similarity, however, underperforms cross-label random selection at larger update magnitudes because the label-wise constraint still limits global diversity and leads to redundancy as more demonstrations are added.

Overall, these results show that while similarity-based selection mitigates some weaknesses of label-wise prompting, the label-wise constraint itself restricts scalability, motivating cross-label selection strategies that allow diversity to grow with update magnitude.

\subsection{Manyshot prompting with Llama-3.1-8B}
\label{app:base_results}

We compare the LLaMA-3.1-8B base model with its instruction-tuned counterpart under identical many-shot prompting conditions on Banking77, using cross-label random selection. As shown in Table~\ref{tab:base_vs_instruct}, the instruction-tuned model substantially outperforms the base model at both update magnitudes, highlighting the importance of alignment for interpreting task instructions and in-context demonstrations.

At low update magnitudes ($n=1$–$5$; not shown), the base model performs near chance, indicating a limited ability to infer task structure from few demonstrations. However, as the number of in-context examples increases, the base model exhibits some gains from $n=50$ to $n=70$ per-class shot, suggesting that many-shot prompting can induce meaningful test-time adaptation even in the absence of instruction tuning. This result indicates that while alignment strongly improves sample efficiency, the benefits of large-scale in-context updates are not exclusive to instruction-tuned models.

\begin{table}[t]%[H]
\centering
\caption{Many-shot prompting for Llama-3.1-8B on Banking 77 (accuracy in \%). Prompts contain $N$ total demonstrations, constructed using $n$ shots per class 
($N = n \times C$, $C{=}77$).}
\label{tab:base_vs_instruct}
\begin{tabular}{c c c}
\toprule
\makecell{Per-class\\shots $n$} & LLaMA-3.1-8B & LLaMA-3.1-8B-Instruct \\
\midrule
50 & 33.9 & 80.8 \\
70 &  35.9 & 82.3 \\
\bottomrule
\end{tabular}
\end{table}

\subsection{Similarity-based retrieval for Dynamic ICL}
\label{app:sim_ret}
For similarity based retrieval, we encode all candidate demonstrations and queries using the sentence-transformers/all-MiniLM-L6-v2 model (384-dimensional embeddings). After generating embeddings, we find n-shot nearest neighbors per class to the query embedding for label-wise similairty selection. We make sure that the retrieved examples don't have the query or same-text duplicates. This ensures there is no leakage. Across all retrieval strategies, we use a fixed prompt template; only the selected demonstrations vary between conditions.

\subsection{Data generation recipe for Reinforced ICL}
\label{app:data_gen}
To generate synthetic reasoning traces for Reinforced ICL on GPQA Diamond shown in Fig.~\ref{fig:ttu_plots}(d), we follow the Chain-of-Thought data generation pipeline described in the Camel AI framework \cite{camelai_datagen}. We instantiate a generator–verifier setup, where both agents are powered by the gpt-oss-120b \cite{openai_gptoss120b} model. The generator produces step-by-step reasoning traces under predefined formatting constraints, and the verifier checks the final answer against the ground-truth label. We retain only examples whose generated final answer exactly matches the correct label, filtering out inconsistent traces. Applying this procedure to the 198 question in GPQA Diamond yields 133 validated reasoning demonstrations, which are used for Reinforced ICL experiments. We used greedy sampling with max tokens set to 12000 to generate CoT traces. We have shown the generator and verifier prompt in Table \ref{tab:generator_prompt} and Table \ref{tab:verifier_prompt} respectively. 

\begin{table}[t]
\centering
\caption{Prompt template used for the Generator agent during data generation for Reinforced ICL.}
\label{tab:generator_prompt}
\footnotesize
\setlength{\tabcolsep}{6pt}
\begin{tabularx}{\linewidth}{p{0.95\linewidth}}
\toprule
\textbf{Generator Prompt} \\
\midrule
Generator agent for creating step-by-step reasoning. Rules: \\
-- Use at most 10 steps. \\
-- No explanations. \\
-- No restating the question. \\
-- End immediately after the answer. \\
-- Do not reference the order of the answers in the question (e.g., option a, b, c). \\
-- For the answer in the \texttt{<a>} tags, ensure it exactly matches one of the listed options. \\
\\
Format: \\
\texttt{<r>} \\
\texttt{1. ...} \\
\texttt{2. ...} \\
\texttt{</r>} \\
\texttt{<a>} \\
\texttt{...} \\
\texttt{</a>} \\
\\
\bottomrule
\end{tabularx}
\end{table}

\begin{table}[t]
\centering
\caption{Prompt template used for the Verifier agent during data generation for Reinforced ICL.}
\label{tab:verifier_prompt}
\footnotesize
\setlength{\tabcolsep}{6pt}
\begin{tabularx}{\linewidth}{p{0.95\linewidth}}
\toprule
\textbf{Verifier Prompt} \\
\midrule
Verified agent for step by step reasoning. \\ For the text within the <a> tags, make sure it matches exactly to the correct answer, word for word.\\
\\
\bottomrule
\end{tabularx}
\end{table}

\subsection{Many-shot prompting on Evaluation Harness}
\label{app:eval_harness}

We report full results across Evaluation Harness tasks with LLaMA-3.3-70B-Instruct in Table \ref{tab:task_success_failure_full} to characterize where many-shot test-time updates succeed and where they fail. These experiments span structured reasoning, and open-ended generation settings, allowing us to examine how the effectiveness of prompt-based updates depends on task structure and information content. The complete results provide finer-grained evidence for the task-dependent patterns discussed in the main paper, and highlight regimes where additional in-context demonstrations introduce useful adaptation signals versus redundant or noisy context. We use BLEU score to evaluate machine translation (WMT16) tasks, F1 score to evaluate reading comprehension (DROP) task and accuracy (in \%) based on exact match for the rest of the tasks.

\begin{table}[H]
\centering
\caption{Task-dependent performance of many-shot test-time updates (LLaMA-3.3-70B-Instruct)}
\label{tab:task_success_failure_full}
\begin{tabular}{l c c c c}
\toprule
Task & Baseline & 4-Shot & 16-Shot & 32-Shot \\
\midrule
\textbf{Domain-Specific / Information-Heavy Tasks} & & & \\
DROP       & 10.8 & 13.1 & 14.2 & 15.8 \\
FDA        & 39.65 & 86.8 & 89.4 & 89.7 \\
SWDE       & 74.17 & 92.9 & 95.1 & 96.4 \\
\midrule
\textbf{Structured Reasoning Tasks (Constrained Output)} & & & \\
ARC-Challenge & 38.82 & 93.72 & 93.48 & 93.45 \\
GSM8K        & 87.56 & 94.7 & 94.8 & 94.8 \\
GPQA (MC)    & 47.99 & 51.9 & 50.0 & 48.8 \\
\midrule
\textbf{Open-Ended Generation Tasks} & & & \\
WMT16 En-De  & 34.51 & 37.0 & 37.4 & 37.5 \\
WMT16 De-En  & 44.76 & 46.3 & 46.9 & 47.0 \\
\bottomrule
\end{tabular}
\end{table}

\subsection{Context growth under many-shot and reinforced test-time updates}
\label{app:context_length_plot}

\begin{figure}[t]
\centering
\includegraphics[width=0.9\linewidth]{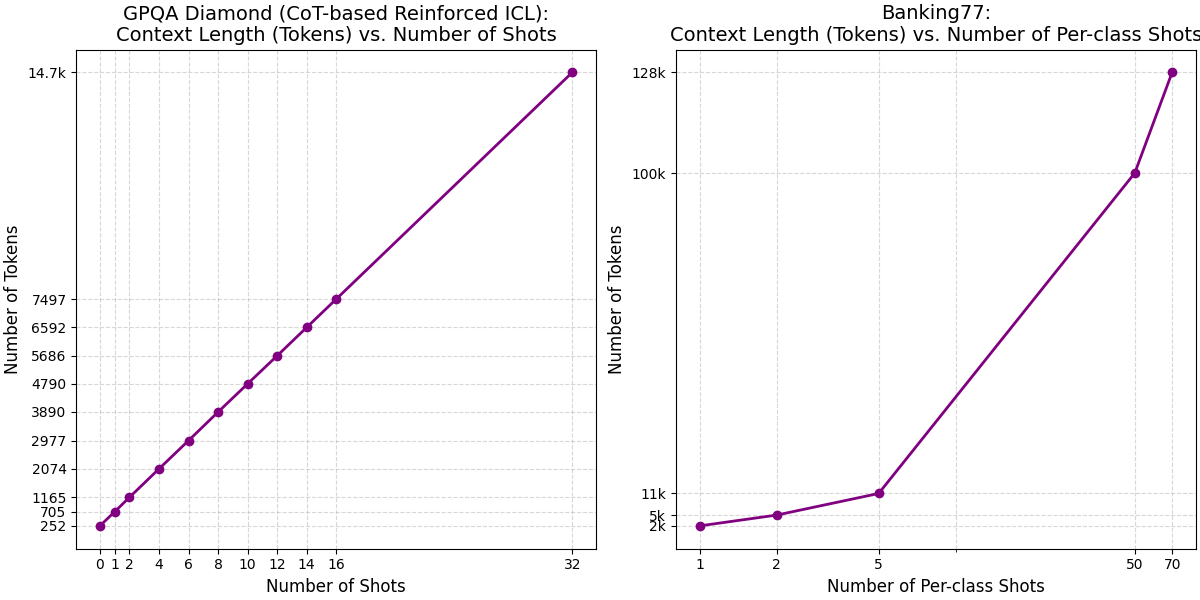}
\caption{Context length scaling with update magnitude}
\label{fig:context_length}
\end{figure}

Figure~\ref{fig:context_length} illustrates how prompt context length (in tokens) scales with the number of demonstrations for two settings: Reinforced ICL on the GPQA Diamond subset (left) and many-shot prompting on Banking77 (right). For Reinforced ICL, the context length grows approximately linearly with the number of reasoning demonstrations included in the prompt, reaching a maximum input length of 14.7k tokens at 32-shot. For Banking77, we vary the number of per-class shots $n$; the total number of demonstrations in the prompt is $N = n \times C$, where $C = 77$ is the number of classes. At $n{=}70$ ($N{=}5390$), the prompt reaches the maximum supported context length of 128k tokens.

\end{document}